\documentclass{article}
\usepackage[top=3cm, bottom=3cm, left=3cm, right=3cm]{geometry}
\usepackage[utf8]{inputenc}
\usepackage{amsmath}
\usepackage{amsfonts}
\usepackage{amsthm}
\usepackage{amssymb}
\usepackage{graphicx}
\usepackage{xcolor}
\usepackage[hidelinks]{hyperref}
\usepackage{caption}
\usepackage{subcaption}
\usepackage{float}
\usepackage{setspace}
\usepackage{fancyhdr}
\usepackage{dsfont}
\usepackage{parskip}
\usepackage{eso-pic}
\usepackage{algorithm}
\usepackage[noend]{algpseudocode}
\usepackage{zref-user}
\usepackage{xcolor}
\usepackage{longtable}
\usepackage{mathtools}
\usepackage{appendix}
\usepackage{cite}
\usepackage{enumitem}
\usepackage{tcolorbox} 
\graphicspath{{../PolyComplet/}}

\usepackage{cite}
\usepackage{textcomp}
\usepackage{wrapfig}
\usepackage{subcaption}

\title{Double-Layer Soft Data Fusion for Indoor Robot WiFi-Visual Localization \footnote{This paper will be submitted to IEEE Sensors Journal}}
\author{Yuehua Ding, Jean-François Dollinger, Vincent Vauchey, Mourad Zghal
\thanks{Yuehua Ding, Jean-François Dollinger, Vincent Vauchey and  Mourad Zghal are with the laboratory CESI LINEACT UR7527, France.}
\thanks{Correspondence: Yuehua Ding, {\tt\small yding@cesi.fr}}
}

\begin{document}

\maketitle

\begin{abstract}
This paper presents a novel WiFi-Visual data fusion method for indoor robot (TIAGO++) localization. This method can use 10 WiFi samples and 4 low-resolution images ($58 \times 58$ in pixels) to localize an indoor robot with an average error distance about 1.32 meters. The experimental test is 3 months after the data collection in a general teaching building, whose WiFi and visual environments are partially changed. This indirectly shows the robustness of the proposed method.   

Instead of neural network design, this paper focuses on a soft data fusion to prevent the unbounded errors in visual localization. A double-layer soft data fusion is proposed. The proposed soft data fusion includes the first-layer WiFi-Visual feature fusion and the second-layer decision vector fusion. Firstly, motivated by the excellent capability of neural network in image processing and recognition, the temporal-spatial features are extracted from WiFi data, these features are represented in image form. Secondly, the WiFi temporal-spatial features in image form and the visual features taken by the robot camera are combined together, and are jointly exploited by a classification neural network to produce a likelihood vector for WiFi-Visual localization. This is called first-layer WiFi-Visual fusion. Similarly, these two types of features can be separately exploited by neural networks to produce another two independent likelihood vectors. Thirdly, the three likelihood vectors are fused by Hadamard product and median filtering to produce the final likelihood vector for localization. This is called the second-layer decision vector fusion. The proposed soft data fusion does not apply any threshold or prioritize any data source over the other in the fusion process. It never excludes the candidate positions of low probabilities, which can avoid the information loss due to a hard decision. The demo video is provided\footnote{The demo video available at: \href{https://youtu.be/\_\_AlHGhDNSI}{https://youtu.be/\_\_AlHGhDNSI}}, and the code will be open to the public after the publication of this work.

\end{abstract}

\section{Introduction}
Mobile robots can bring enormous social and economic benefits in
dealing with the challenges brought by the aging society and labor shortage. Its applications cover a wide range of areas, such as healthcare and hospitals, industry, warehousing, logistics, transportation, laboratories etc. \cite{Umetani}. 
Mobile robots can replace manual tasks such as transporting goods, surveillance patrols, operations in dangerous environments and repetitive work. The high accuracy of indoor positioning is the premise to ensure that the mobile robot can complete tasks autonomously  \cite{Lee}.

The problem of positioning in external environment can be solved by the ubiquitous global navigation satellite system (GNSS). However, GNSS has some difficulties in indoor environment due to complex indoor conditions, such as signal attenuation, multi-path effect, non line of sight (NLOS) problems. On the other hand, indoor positioning accuracy, generally, needs to be at meter/sub-meter level to ensure that a robot can take a correct route.

To meet this challenge, many indoor positioning technologies have been proposed. In general, we have two categories of methods for indoor positioning: visual methods and radio methods. Visual methods are based on computer vision, image processing and artificial intelligence (AI). Radio methods are based on radio signal processing and the geometric principle or adaptation of radio data. These two types of technologies have been applied to the positioning of mobile robots, significantly increasing robots' competence. Currently, there is no technology as dominant in the field of indoor positioning as GNSS in the field of outdoor positioning. Each positioning technology has significant advantages and obvious disadvantages. The road to a true autonomous robot is still long.

Visual methods \cite{Radwan} \cite{Valada} provide cost-effective, detailed and context-rich positioning/maps. Its image data is very suitable for neural networks processing, which achieves a great success in image recognition. However, this method requires a lot of labeled images to achieve good precision, which requires significant calculation time. Additionally, they tend to fail in homogeneous spaces with limited functionality such as long corridors or poor interior lighting conditions. One interesting example is that the glass door looks different during the day and at night. One can see the outdoor scenery from the glass door during the day, but at night the glass door reflects the indoor scene. This visual difference brought by the changes of luminosity condition can significantly degrade the localization performance.

The radio methods are quite rich. There exists various technologies such as ultra wide band (UWB) \cite{Bach}, frequency modulated continuous wave (FMCW) \cite{Venon}, WiFi\cite{Jun}-\cite{Chen} etc. UWB can achieve high accuracy localization, but it has a risk of spectral conflict, additional sensor deployment is needed. UWB based localization requires the implementation of a specific anchor architecture, its localization performance is related to the density of sensor deployment. FMCW can combine radar technology and laser technology to achieve an accurate localization, however, it is very expensive in cost, and it is complex in hardware and software development. WiFi is already ubiquitously deployed with a huge number of access points (AP), it is cheap and convenient, but its accuracy is not high. Compared to visual methods, signal processing for radio methods is not very complicated. In addition, light detection and ranging (LiDAR) \cite{VV} can provide long-range, high-resolution sensing, but it is very expensive and still suffers from limitations in the case of similar geometric environments and robot kidnapping (with unknown initial position).

Based on the previous comparison, the authors of this paper observe that both camera and WiFi are ubiquitously available with low cost. Localization using vision can achieve excellent accuracy in most of the time, however, it may produce catastrophic unbounded localization errors between the points having similar visual contexts regardless their physical distance. WiFi localization is not so accurate as the former due to significant fluctuations of WiFi signal strength. Thanks to the propagation properties of radio signal, WiFi's coverage range prevents its positioning error from being a catastrophic value.

Inspired by the complementary natures of WiFi localization and visual localization, this paper focuses on a WiFi-Visual data fusion, instead of neural network design. The objective is to combine the generally excellent localization accuracy of visual positioning and the bounded localization errors of WiFi localization. The contributions of this paper are summarized as follows:

Firstly, instead of exploiting WiFi signal strength directly, intrinsic features are extracted from WiFi data. These features reflect the temporal-spatial spectrum of different WiFi access points, and the correlation across different access points as well. Compared with the fluctuating signal strength, spectrum and correlation properties are more stable signatures of a position. They are represented in the form of image to adapt the strong capability of neural network in image recognition.

Secondly, a double-layer soft data fusion is proposed. In the process of data fusion, neither threshold nor data source priority is applied. In the first-layer fusion, the WiFi features in image form and the visual features provided by the camera are combined together, and jointly exploited by a neural network. By modeling the localization problem as a classification, a likelihood vector is produced by the neural network as a response to the input features. In the second layer fusion, another two likelihood vectors are obtained by separately exploiting the WiFi features and visual features. The decision vector fusion is realized by Hadamard product and median filtering on the three likelihood vectors to produce the final decision vector for localization. 

Thirdly, in-field experiments with a true robot TIAGO++ are carried out in quite a general scenario at the ground floor of a teaching building. The test experiment shows that the proposed localization method can localize the robot with an average error distance about 1.32 meters by using 10 WiFi samples and 4 low-resolution images. In particular, the test experiment is 3 month after the data collection, which reflects the robustness of the proposed method.

The rest of this paper is organized as follows: Section II presents the related works; Section III describes a general model for WiFi-Visual robot localization. Section IV proposes a WiFi-Visual data fusion method for localization. The experiment results are discussed in Section V. Section VI draws the conclusions.

\textit{Notations:}
Capital letters of boldface are used for denoting
matrices. Lowercase letters of boldface denote column vectors. $(\cdot)^T$ denotes the operation of matrix transpose. $\bigodot $ represents Hadamard product. $\mathbb{A}({\bf M})$ (or $\mathbb{P}({\bf M})$) represents the matrix obtained by element-wise operation of taking the amplitudes (or phases) of the elements in matrix ${\bf M}$. $\| (\cdot)\|_2$ takes the $L_2$-norm of $(\cdot)$. $\mathbb{E}[\cdot]$ is the statistical expectation. 

\section{Related works}
\subsection{WiFi localization}
WiFi draws increasing attentions in the domain of indoor robot localization. Its omnipresence can compensate the absence (weak presence) of GPS signal in indoor environments. Its low-cost deployment makes its wide application economically possible. However, in technique aspect, WiFi signal strength fluctuation can significantly degrade the localization accuracy. To improve the accuracy, denser fingerprint sampling can be adopted to achieve higher accuracy \cite{Jun}. One should note that denser fingerprint sampling need more labor cost and time cost. \cite{Sun} uses data augmentation technique to enlarge the WiFi fingerprint dataset for higher localization accuracy. Actually, we can not infinitely improve the accuracy by increasing the fingerprint sampling density due to the ambiguity caused by sensitivity level of WiFi device. WiFi signal strength fluctuation can largely increase the surface of ambiguity zone. To deal with this issue, other signal properties such as direction of arrival (DoA), time of arrival (ToA), time difference of arrival (TDoA), time of flight (ToF) are exploited \cite{Ayyalasomayajula}. \cite{Ayyalasomayajula} constructs a heat map by using these radio properties, and a machine learning method is employed for localization. \cite{Ayyalasomayajula1} \cite{Chen} \cite{Hoang} exploit the WiFi channel state information (CSI) for localization by using a special WiFi card whose CSI is available for users. The methods\cite{Ayyalasomayajula} \cite{Ayyalasomayajula1} \cite{Chen} exploiting additional radio properties, such as DoA, ToA, TDoA, ToF or CSI, can achieve very good localization performance, but they need additional or special devices, which limits their wide applications. 
\subsection{WiFi-Visual localization}
The rich environment information in image motivates the research in visual localization. Its localization accuracy and low-cost feature make its wide application possible. Nevertheless, the computing burden for image matching, and the visual aliasing \cite{Bouazzaoui} \cite{Zhao} in homogeneous environments are two main challenges. To cope with these challenges, WiFi-Visual localization is a good balance. \cite{Ayyalasomayajula} realizes a WiFi localization in the way of image processing, which represents the radio properties of WiFi signal in image form to adapt the strong competence of neural network in image processing. A data fusion approach based on threshold is proposed by \cite{Redzic} to integrate image and WiFi. \cite{Huang} proposes a multi-scale strategy for WiFi-Visual localization. An indoor localization based on sequential data fusion is proposed by \cite{Almansoub}. According to \cite{Almansoub}, WiFi signals are used for coarse localization, and the images are exploited to refine the localization results, which is also the case in \cite{Tang1}. In \cite{Tang}, a sequential-multi-decision fusion is proposed for WiFi-Visual localization, where Gaussian process regression and hybrid whale optimization algorithm are used.  

\section{System model}
\begin{figure}[h]
\centering
\includegraphics[width=5in]{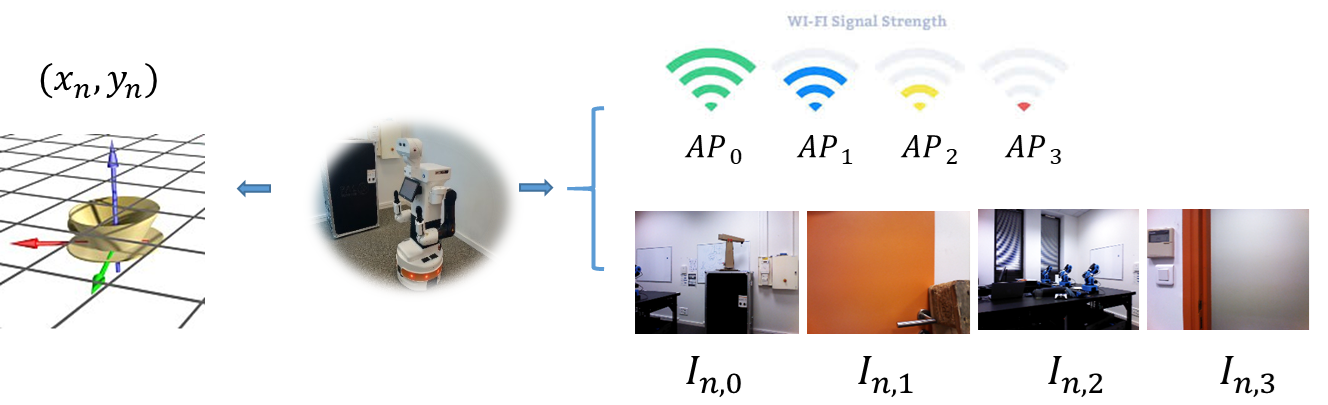} 
\caption{WiFi-Visual robot localization system.}
\label{system}
\end{figure}

A WiFi-Visual robot localization system is illustrated by Fig. \ref{system}. 
A robot localizes itself by processing the data collected by its WiFi antenna and camera. Mathematically, the positioning process can be abstracted by a function as follows:
\begin{equation}
 ({\hat x}_n,{\hat y}_n) = f_{wv}({\bf S}_n, {\bf I}_n)   \label{prediction model}
\end{equation}
where $f_{wv}(\cdot)$ represents a WiFi-Visual localization method, such as but not limited to the methods based on KNN, LSTM or CNN. $({\hat x}_n,{\hat y}_n)$ represents the estimated coordinate of the $n^{th}$ position $({x}_n,{y}_n)$, where WiFi data ${\bf S}_n$ and image data ${\bf I}_n$ are collected for localization. The WiFi data ${\bf S}_n$ is usually a group of received signal strength indicator (RSSI) samples, ${\bf S}_n$ contains $M$ RSSI samples of $K$ access points, ${\bf S}_n$ can be expressed as an $M \times K$ matrix in Eq. (\ref{WiFi data}) 
\begin{equation}
 {\bf S}_n = \begin{bmatrix}
s_{n,0,0} & s_{n,0,1} &\cdots& s_{n,0,K-1}\\
s_{n,1,0} & s_{n,1,1} &\cdots& s_{n,1,K-1}\\
\cdots & \cdots &\cdots& \cdots\\
s_{n,M-1,0} & s_{n,M-1,1} &\cdots& s_{n,M-1,K-1}\\
\end{bmatrix}  \label{WiFi data}
\end{equation} 
The image data ${\bf I}_n$ contains a sequence of $S$ images taken by the camera of the robot at position $({x}_n,{y}_n)$ . ${\bf I}_n$ can be represented as follows:
\begin{equation}
 {\bf I}_n = \begin{bmatrix}
{\bf I}_{n,0} & {\bf I}_{n,1} &\cdots& {\bf I}_{n,S-1}
\end{bmatrix}  \label{Image data}
\end{equation} 
The objective is the minimization of localization error.
\begin{equation}
{min\rm{~~}}\| ({\hat x}_n,{\hat y}_n) - ({x}_n,{y}_n) \|_2 \label{objective}
\end{equation} 
In particular, the robot can also perform the WiFi localization or image localization separately, which can be respectively formulated as follows :
\begin{equation}
 ({\bar x}_n,{\bar y}_n) = f_w({\bf S}_n)  \label{wifi prediction} 
\end{equation}
\begin{equation}
 ({\tilde x}_n,{\tilde y}_n) = f_v({\bf I}_n) \label{visual prediction}   
\end{equation}
where $f_{w}(\cdot)$ and $f_{v}(\cdot)$ represent a WiFi-only localization method and an image-only localization method, respectively.

In this paper, the localization is realized by means of classification. The plan is partitioned into $N$ small pieces of disjoint areas, the $n^{th}$ piece is defined as class $L_n$ with $n=0,1,\cdots, N-1$. The localization in (\ref{prediction model}) is re-modeled as:
\begin{equation}
 ({\hat x}_n,{\hat y}_n) \leftarrow \hat{L}_n = f_{wv}({\bf S}_n, {\bf I}_n)  \label{classification}
\end{equation}
In (\ref{classification}), $({\hat x}_n,{\hat y}_n)$ is the centre position of the area labelled as class $\hat{L}_n$. $\hat{L}_n$ is the estimate of ${L}_n$

\section{Proposed method}
This section presents a novel robot localization method, which jointly exploits the WiFi and visual perception. In particular, the information provided by WiFi perception can be independently used for WiFi localization. It is also the case for visual perception in visual localization. The exploitation of different types of information is discussed as follows, one should note that the preprocessing on ${\bf S}_n$ is needed before the real processing. With a little abuse of notation, ${\bf S}_n$ is still used to represent the preprocessed data.

\subsection{WiFi Localization}
The main challenge of WiFi localization is the significant fluctuation of WiFi signals' strength in time domain. To cope with this problem, the intrinsic features of WiFi signals are extracted from the WiFi data in Eq. (\ref{WiFi data}). These features include the temporal-spatial spectrum, the correlation matrix of access points.

\subsubsection{Temporal-spatial spectrum}

A row in Eq. (\ref{WiFi data}) is a WiFi RSSI sample from different access points, while a column represents a sequence measuring the temporal variation of RSSI from a single access point. To characterize the features, two dimensional Discrete Fourier Transform (DFT) can be applied on ${\bf S}_n$.
\begin{equation}
\Tilde{\bf S}_n = {\bf F}^T_M{\bf S}_n{\bf F}_K  \label{DFT}
\end{equation}
where ${\bf F}_M$ (or ${\bf F}_K$) represents the DFT matrix. Without loss of generality, ${\bf F}_M$ is given by
\begin{equation}
    {\bf F}_M = \frac{1}{\sqrt{M}} \begin{bmatrix}
1&1&1&\cdots &1 \\
1&\omega&\omega^2&\cdots&\omega^{M-1} \\
1&\omega^2&\omega^4&\cdots&\omega^{2(M-1)}\\ 1&\omega^3&\omega^6&\cdots&\omega^{3(M-1)}\\
\vdots&\vdots&\vdots&\ddots&\vdots\\
1&\omega^{M-1}&\omega^{2(M-1)}&\cdots&\omega^{(M-1)(M-1)}
\end{bmatrix} \label{DFT matrix}
\end{equation}
with $\omega = e^{-2\pi j/N}$, $j^2 = -1$.

$\Tilde{\bf S}_n$ represents the temporal-spatial spectrum of the WiFi signal strength received at the $n^{th}$ position. The $m^{th}$ row of $\Tilde{\bf S}_n$ represents the spatial spectrum with respect to different access points at the instant of the $m^{th}$ sampling.  The $k^{th}$ column represents the temporal spectrum of the $k^{th}$ access point.

To exploit the temporal-spatial spectrum for localization, the amplitude and phase of $\Tilde{\bf S}_n$ are extracted as two features, represented by $\mathbb{A}(\Tilde{\bf S}_n)$ and $\mathbb{P}(\Tilde{\bf S}_n)$ respectively.

\subsubsection{Correlation matrix of access points}
The correlation between access point $k$ and ${k'}$ is theoretically given by
\begin{equation}
r_{n,k,{k'}} = \frac{\mathbb{E}[(s_{n,m,k}-\mu_k) (s_{n,m,{k'}}-\mu_{k'})]}{\sigma_k\sigma_{k'}}
\end{equation}
where $\mu_k$ and $\sigma_k$ are the mean and standard deviation of RSSI from access point $k$, respectively. For the preprocessed ${\bf S}_n$, $\mu_k$ is centered to 0, and $\sigma_k$ is normalized to 1. In practice, $\mathbb{E}[\cdot]$ is replaced by averaging. The correlation matrix of access points at $n^{th}$ position is calculated as:
\begin{equation}
{\bf R}_n = \frac{{\bf S}_n^T{\bf S}_n}{M} \label{correlation matrix}
\end{equation}
In (\ref{correlation matrix}), ${\bf R}_n$ is a $K \times K$ symmetric matrix.

\begin{figure}[h]
\centering
\includegraphics[width=3.5in]{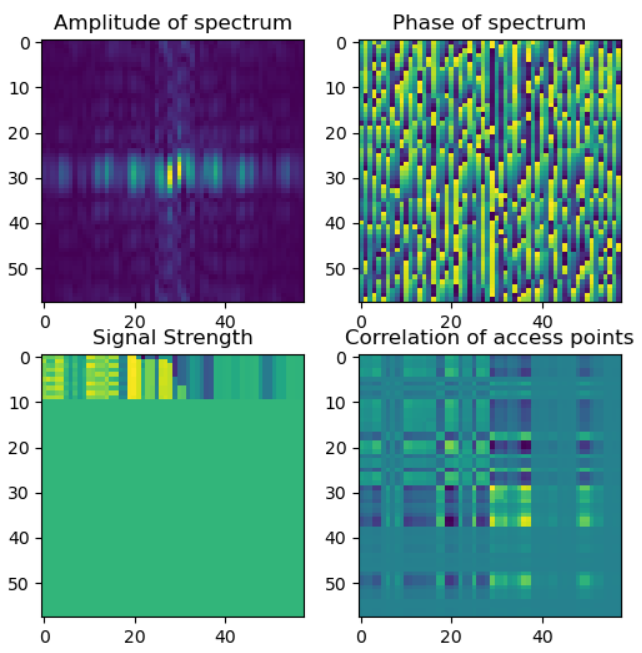} 
\caption{WiFi RSSI features in image form}
\label{WiFi}
\end{figure}

In WiFi perception, $\mathbb{A}(\Tilde{\bf S}_n)$, $\mathbb{P}(\Tilde{\bf S}_n)$ and ${\bf R}_n$  are three different features  to be used. To keep the three matrices of the same dimensions, additional points can be added to ${\bf S}_n$ or ${\bf R}_n$ if $ M \neq K$. Fig.\ref{WiFi} is an example of the visualization of WiFi features.

According to (\ref{wifi prediction}), the image-form WiFi features can be separately used as follows: 
\begin{equation}
 {\bf p}_w = f_{w}(\mathbb{A}(\Tilde{\bf S}_n), \mathbb{P}(\Tilde{\bf S}_n), {\bf R}_n,{\bf S}_n )   \label{likelihood_wifi}
\end{equation}
where $f_{w}$ represents a neural network in this paper. ${\bf p}_w$ is a vector of conditional probability given by
\begin{equation}
{\bf p}_w = [p((x_1,y_1)|{\bf S}_n), p((x_2,y_2)|{\bf S}_n),\cdots, p((x_N,y_N)|{\bf S}_n) ]^T
\end{equation}
Its element $p(x_i,y_i|{\bf S}_n)$ indicates the likelihood of the $i^{th}$ candidate position. For WiFi localization, the estimated position $({\bar x}_n,{\bar y}_n)$ is given by 
\begin{equation}
p(({\bar x}_n,{\bar y}_n)|{\bf S}_n) = max {\rm{~~}} p((x_i,y_i)|{\bf S}_n) \label{likelihood maximization wifi} \end{equation}
\subsection{Visual Localization}
In this paper, the visual perception is relatively simple. The photo sequence ${\bf I}_{n,0}, {\bf I}_{n,1},\cdots {\bf I}_{n,S-1}$ taken at the $n^{th}$ position are sent to the neural network $f_v(\cdot)$ in (\ref{visual prediction}) for learning in the training process or for localization in test. 
\begin{equation}
 {\bf p}_v = f_{v}({\bf I}_n)   \label{likelihood_vision}
\end{equation}
where ${\bf p}_v$ is a probability vector produced by the neural network $f_v(\cdot)$. ${\bf p}_v$ is expressed by
\begin{equation}
{\bf p}_v = [p((x_1,y_1)|{\bf I}_n), p((x_2,y_2)|{\bf I}_n),\cdots, p((x_N,y_N)|{\bf I}_n) ]^T
\end{equation}
For the localization by vision, the coordinates corresponding to the maximum element in Eq. (\ref{likelihood_vision}) is taken as the robot position, as follows
\begin{equation}
p(({\tilde x}_n,{\tilde y}_n)|{\bf I}_n) = max {\rm{~~}} p((x_i,y_i)|{\bf I}_n) \label{likelihood maximization visual} \end{equation}
\subsection{First-Layer WiFi-Visual Feature Fusion}
Both WiFi features and image features can be represented in the image form. To facilitate the data fusion, it is necessary to harmonize their dimensions, in particular, ${\bf R}_n$, ${\bf S}_n$, ${\bf I}_n$ are reshaped as the pictures of the same dimensions. They are jointly exploited to localize the robot as follows:
\begin{equation}
 {\bf p}_{wv} = f_{wv}(\mathbb{A}(\Tilde{\bf S}_n), \mathbb{P}(\Tilde{\bf S}_n), {\bf R}_n,{\bf S}_n , {\bf I}_n) \label{likelihood_wifi_vision}  
\end{equation}
where ${\bf p}_{wv}$ is a probability vector produced by the neural network $f_{wv}$. ${\bf p}_{wv}$ is expressed by
\begin{equation}
{\bf p}_{wv} = [p((x_1,y_1)|{\bf S}_n,{\bf I}_n),\cdots, p((x_N,y_N)|{\bf S}_n,{\bf I}_n) ]^T
\end{equation}
Eq. (\ref{likelihood_wifi_vision}) can provide a position estimate of the robot. The position estimate is given by
\begin{equation}
p(({\underline {x}}_n,{\underline y}_n)|{\bf S}_n, {\bf I}_n) = max {\rm{~~}} p((x_i,y_i)|{\bf S}_n,{\bf I}_n) \label{likelihood maximization fusion} \end{equation}
\subsection{Second-Layer Decision Vector Fusion}
For the data fusion of WiFi data and image data, we neither prioritize different data sources nor use the prediction based on one data source to restrict the prediction based on other data source. A natural data fusion is adopted with respect to the likelihoods provided by the predictions from different data sources. To fuse the information in ${\bf p}_{w}$, ${\bf p}_{v}$, ${\bf p}_{wv}$, the Hadamard product of ${\bf p}_{w}$, ${\bf p}_{v}$, ${\bf p}_{wv}$ is taken as a new vector of likelihood. 
\begin{equation}
 {\bf p}_{wvm} = {\bf p}_{w} \bigodot {\bf p}_{v} \bigodot {\bf p}_{wv} \label{likelihood_wifi_vision_mask}
\end{equation}
One should note that the sum of ${\bf p}_{wvm}$ given by (\ref{likelihood_wifi_vision_mask}) is not necessarily equal to 1. ${\bf p}_{wvm}$ should be normalized to a regular probability vector. The principle of Eq.(\ref{likelihood maximization fusion}) can also be applied on  Eq. (\ref{likelihood_wifi_vision_mask}) to localize the robot. However, in this paper, an additional median filtering is applied on ${\bf p}_{wvm}$, ${\bf p}_{w}$ and ${\bf p}_{v}$ to produce the final likelihood vector.  
\section{Experiments}

\subsection{Experiment platform}
The experiment platform is a real physical robot TIAGO++ \cite{Pal}. TIAGO++ is a fully ROS-based, customisable robot platform, it is adapt to the research needs in AI, machine learning, human-robot interaction (HRI), and manipulation. WiFi card of 802.11ax WiFi 6 and RGB-D camera are integrated into the robot platform, they can be used directly to collect WiFi and visual data. For performance evaluation, thanks to the LiDAR position and mapping system in TIAGO++, simultaneous localization and mapping (SLAM) can be performed in the experiments to provide a reliable map and real-time positions with centimeter-level accuracy, which are considered as ground-truth values in performance evaluation in the following experiments.

\begin{figure}[h]
\centering
\includegraphics[width=5 in]{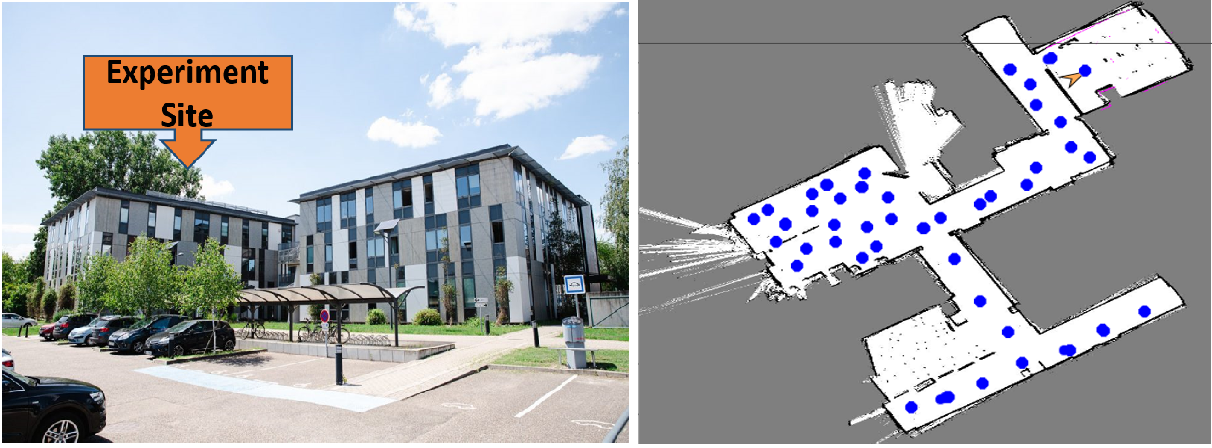} 
\caption{Mapping in the ground floor of the teaching building.}
\label{Mapping}
\end{figure}

To examine the effectiveness of the proposed method, extensive experiments are carried out, including mapping, data collecting and testing. To keep the generality of the experimental environment, the experiments are carried out in the ground floor of a teaching building with usual activities, as shown in Fig. \ref{Mapping}. The training process is completed in the server. In the experiment, $K = 58$ logic access points of WiFi are considered, $M = 10$ WiFi RSSI samples and $S= 4$ photos are taken for a one-time localization. The performance is evaluated in terms of root mean square error (RMSE), mean absolute error (MAE), and standard deviation (STD), respectively. They are computed as follows:
\begin{equation}
RMSE = \sqrt{\frac{1}{N_{test}}\sum_{i=0}^{N_{test}-1} [({\hat x}_i-{x}_i)^2 + ({\hat y}_i-{y}_i)^2]}
\end{equation} 
\begin{equation}
MAE =\frac{1}{N_{test}} {\sum_{i=0}^{N_{test}-1} (|{\hat x}_i-{x}_i| + |{\hat y}_i-{y}_i|)}
\end{equation} 
\begin{equation}
STD = \sqrt{\frac{1}{N_{test}}\sum_{i=0}^{N_{test}-1} ({e_i}-\bar{e})^2 }
\end{equation} 
where $N_{test}$ represents the number of test points, 
$({\hat x}_i,{\hat y}_i)$ is the estimated coordinate of test point $i$, $({x}_i,{y}_i)$ is the ground-truth value, which can be provided by the LiDAR system of the platform, ${e_i}$ is the error of $i^{th}$ point of localization, $\bar{e}$ is the average localization error.

\subsection{Data collecting}
\begin{figure}[h]
\centering
\includegraphics[width=3.5in]{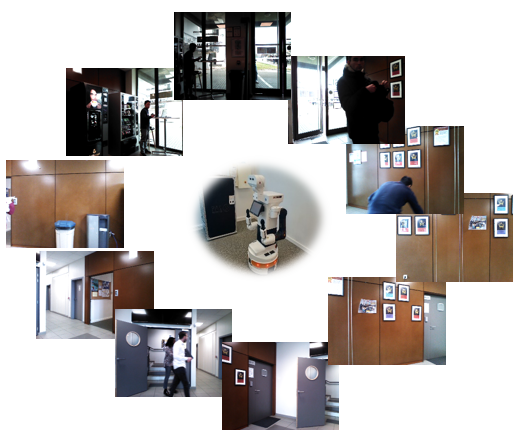} 
\caption{Visual perception.}
\label{Visual}
\end{figure}

\begin{figure}[h]
\centering
\includegraphics[width=5in]{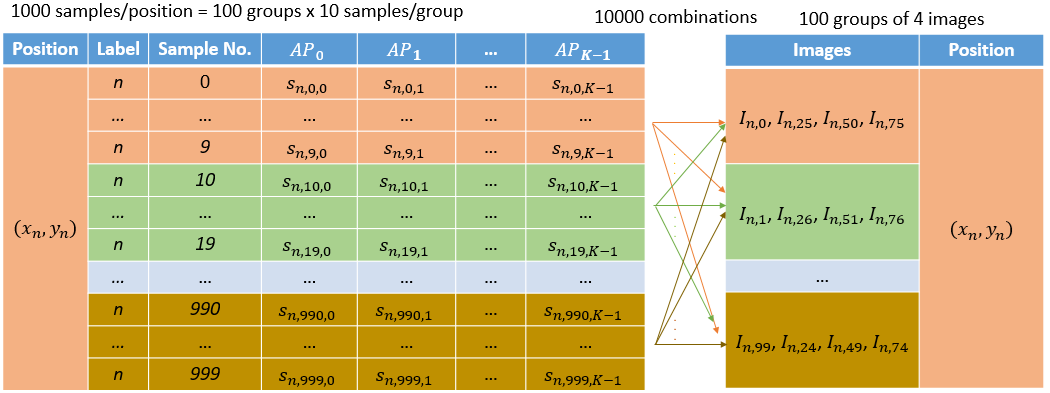} 
\caption{Data association}
\label{data association}
\end{figure}

The sampled positions of data collection are marked in blue color in the map of Fig.\ref{Mapping}. For each sampled position, TIAGO++ takes 1000 WiFi RSSI samples and 100 photos around itself. In the image collecting process, TIAGO++ rotates itself by 3.6 degrees after taking each photo, as illustrated by Fig. \ref{Visual}. 
For a given sampled position, its 1000 WiFi samples are divided into 100 groups with 10 samples in each group. Its 100 photos are organized into 100 groups with 4 photos in each group. The 4 photos in a group should be uniformly spaced in terms of the camera shooting angle. Two neighboring groups are offset by a shooting angle difference of 3.6 degrees, as indicated by Fig. \ref{data association}. To associate the WiFi data and the image data, all possible combinations between 100 WiFi groups and 100 images groups are taken to construct 10000 training WiFi-Visual samples. 
\subsection{Training}
\begin{figure}[h]
\centering
\includegraphics[width=4in]{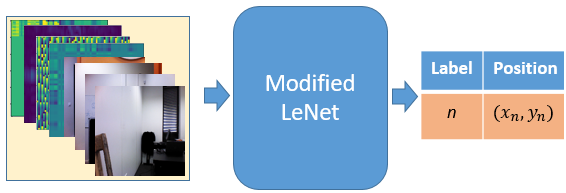} 
\caption{Training}
\label{Training}
\end{figure}

In the training process, we considered four different artificial neural networks (ANN) structures (ResNet \cite{ResNet}, LeNet \cite{LeNet}, LSTM \cite{LSTM}, VLocNet \cite{VLocNet1} \cite{VLocNet2}) to adapter our dataset. Tab. \ref{Network Comparison} provides a comparison among these structures. ResNet is an excellent structure for image recognition, however, it is easy to have overfitting with our dataset, VLocNet is also based on ResNet. We have no overfitting with LSTM and LeNet, it is not easy to harmonize the WiFi feature and image feature to adapt the 
temporal sequential characteristics of LSTM. Hence, we choose LeNet as a basic structure for training. Fig. \ref{Training} illustrates the framework of training with the combinations of WiFi features and visual features. For the training with WiFi features (or image features) only, the image features (or WiFi features) can be removed directly.  
\begin{table}[h]
\centering
\caption{Network Comparison}
\label{Network Comparison}
\setlength{\tabcolsep}{3pt}
\begin{tabular}{|p{70pt}|p{40pt}|p{40pt}|p{40pt}|p{40pt}|}
\hline
  &ResNet&LeNet&LSTM&VLocNet \\
\hline
Overfitting & yes& no& no & yes  \\
\hline
Image adaptive & yes&yes&no &yes  \\
\hline
Complexity & high&low &low &high  \\
\hline
\end{tabular}
\label{tab1}
\end{table}

\subsection{In-field Test}
To examine the performance the proposed algorithm, the in-field test is carried out with TIAGO++ moving in the teaching building. It is interesting to note that the test and data collection are temporally separated by 3 months. During this period, the WiFi infrastructures are partially reorganized by the technique staff, the layout of the teaching building has been partially changed to adapt to the students' various activities. The changing environment makes the localization more complicated, but also increases the robustness of the method.

\begin{figure}[h]
\centering
\includegraphics[width=3.5in]{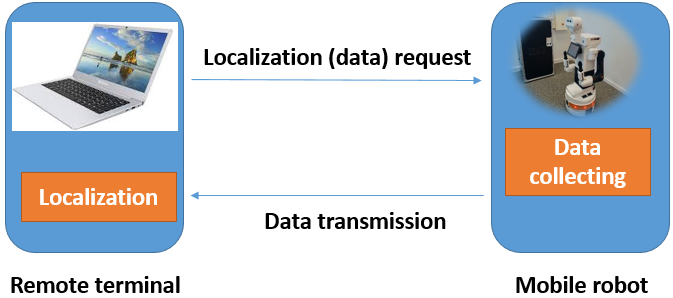} 
\caption{The framework of in-field test}
\label{Test framework}
\end{figure}

In the test, we choose remote interactions between a remote terminal (PC) and the robot to simulate the working environment and reduce interference, however, we don't deliberately change the natural distribution of the people around the robot in the building. Fig. \ref{Test framework} shows the test framework, a remote PC and the robot TIAGO++ are connected to the same WiFi, a localization request is sent from the remote PC to the robot, the robot collects $M = 10$ WiFi RSSI samples and $S= 4$ images, and sends them to the remote PC. The remote PC can use the received data and 
the trained model to localize the robot.
\subsection{Performance comparison before and after soft data fusion}
\begin{figure}[h]
\centering
\includegraphics[width=5in]{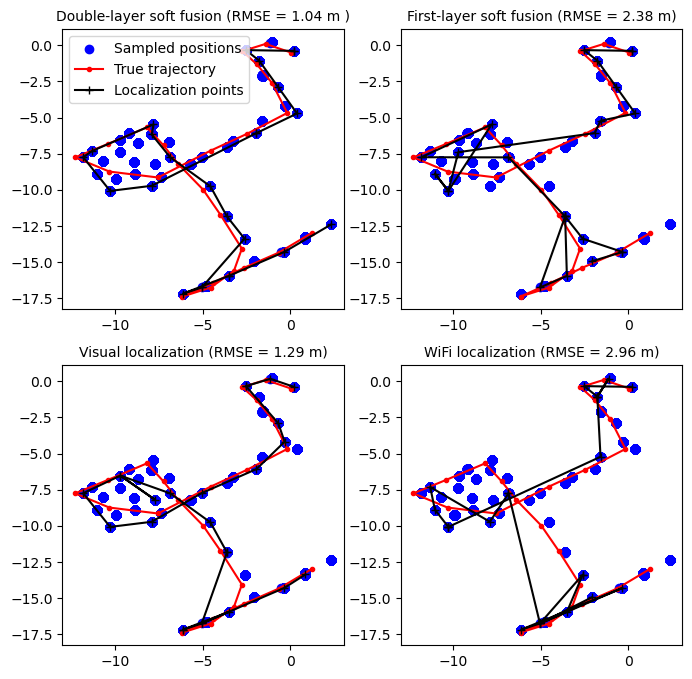} 
\caption{Trajectory comparison among different localization modes}
\label{Trajectory comparison}
\end{figure}

In one of the tests, the robot moves along a trajectory connecting halls and corridors, its positions are estimated by the remote terminal. The positions provided by LiDAR can be considered as the ground truth positions, as shown by the red trajectory in Fig. \ref{Trajectory comparison}. 

\begin{figure}[h]
\centering
\includegraphics[width=5in]{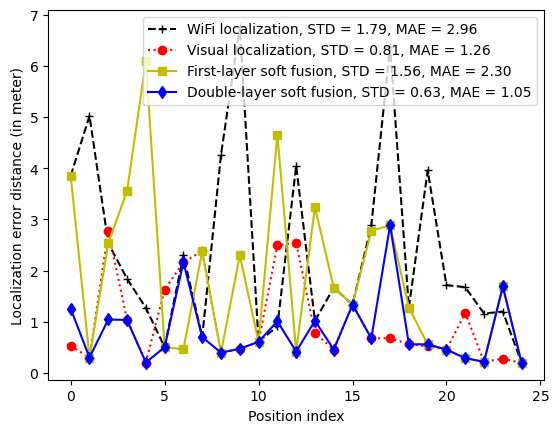} 
\caption{Localization error comparison}
\label{Error comparison}
\end{figure}

\begin{figure}[h]
\centering
\includegraphics[width=5in]{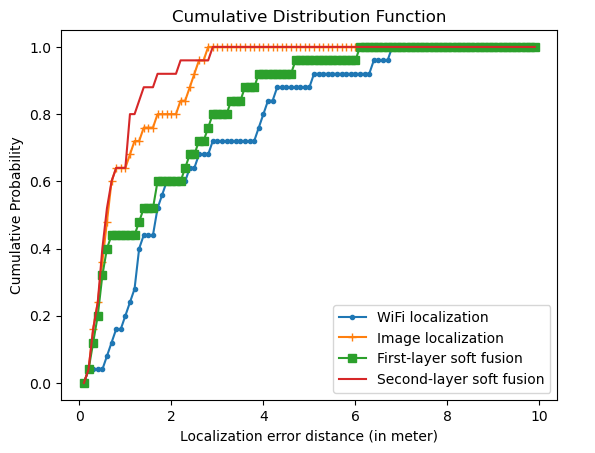} 
\caption{Localization error comparison (CDF)}
\label{CDF comparison}
\end{figure}

Fig. \ref{Trajectory comparison} compares the localization performances of the WiFi localization realized by Eq.(\ref{likelihood_wifi}), the visual localization by Eq.(\ref{likelihood_vision}), the localization provided the first-layer soft fusion by Eq.(\ref{likelihood_wifi_vision}) and the double-layer soft fusion localization after meidan filtering. 
We can observe that the localization RMSE of the first-layer soft fusion is 2.38 m, it is between WiFi localization (RMSE = 2.96 m) and Visual localization (RMSE = 1.29 m). The double-layer soft fusion has the best RMSE accuracy (RMSE = 1.04 m), its trajectory can closely follow the true trajectory. Detailed comparison is quantified by Fig. \ref{Error comparison} and Fig. \ref{CDF comparison}.

Fig. \ref{Error comparison} shows the localization error distance of each estimated position. The maximum error of the double-layer soft fusion is about 3 meters, which is the same case for the visual localization. One should note that the occurrence of big-error localization of double-layer soft fusion
is much less than the visual localization. For double-layer soft fusion, most of its localization errors are smaller than 2 meters. Fig. \ref{CDF comparison} shows the statistics of the localization accuracy in terms of cumulative density function (CDF). About 90\% of the localization points by double-layer soft fusion can achieve an accuracy less than 2 meters, this percentage is only 80\% for the visual localization. In addition, double-layer soft fusion localization has better performance than the others in terms of MAE and STD, as shown in Fig. \ref{Error comparison}.

\subsection{Comparison with other data fusion methods}
{\color{black}The double-layer soft data fusion method is compared with the following data fusion criteria:

$\cdot$ Data fusion based on a threshold of distance $d$: the threshold is a value of distance. A circle of a given radius (threshold) is chosen, the circle centre is given by WIFI localization. The WiFi-Visual localization will completely or partially restricted by this circle.

$\cdot$ Data fusion based on a threshold of probability $\gamma$: the threshold is a probability value. The candidates are filtered according to their individual probabilities, or the sum of their probabilities, in stead of their physical distances to a reference point.

$\cdot$ Top-$K$ data fusion: $K$ is a fixed number. $K$ best candidates are chosen for localization refinement. 

These data fusion criteria or their hybrid versions are widely used in the existing literature \cite{Redzic}-\cite{Tang}.

Another group of tests with $N_{test} = 98$ is done to compare the proposed method with the above data fusion strategies. The data fusion idea of the very recent method in \cite{Tang} (named 'Tang's method') is also compared with the proposed method.

Fig.\ref{CDF comparison-d} shows the CDF performances of the proposed method and the distance-threshold data fusion. One can note that the distance-threshold data fusion can refine the localization well when the localization error is smaller than 2 meters, in this case, it performance is a little bit better than the proposed method. However, when the localization error increases, it is not so effective to refine the accuracy. Even if a bigger threshold value such as $d= 6$ m is used, there is little performance improvement. Actually, increasing the threshold value means relaxing the constraint. Fig.\ref{CDF comparison-d} shows that the proposed method is more effective to limit the large errors.

\begin{figure*}[htb]
    \centering
   
    \begin{subfigure}[t]{0.49\textwidth}
        \includegraphics[clip, width=\textwidth]{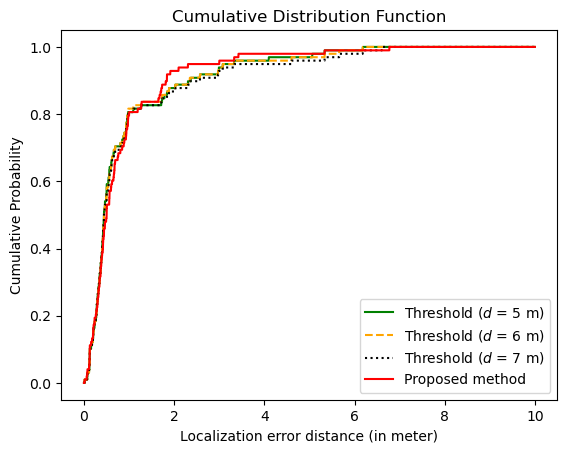}
        \caption{CDF comparison with distance-threshold method}
        \label{CDF comparison-d}
    \end{subfigure}
    \begin{subfigure}[t]{0.49\textwidth}
        \includegraphics[clip, width=\textwidth]{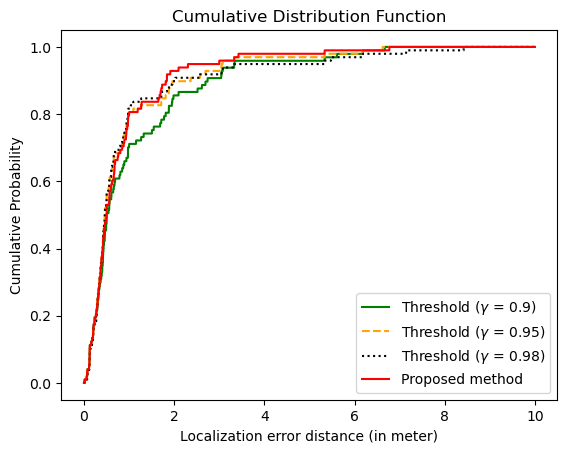}
        \caption{CDF comparison with probability-threshold method}
        \label{CDF comparison-gamma}
    \end{subfigure}
    \begin{subfigure}[t]{0.49\textwidth}
        \includegraphics[clip, width=\textwidth]{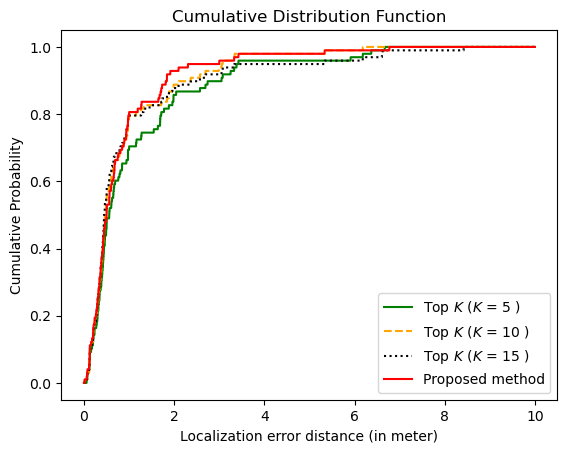}
        \caption{CDF comparison with Top-$K$ method}
        \label{CDF comparison-K}
    \end{subfigure}
    \begin{subfigure}[t]{0.49\textwidth}
        \includegraphics[clip, width=\linewidth]{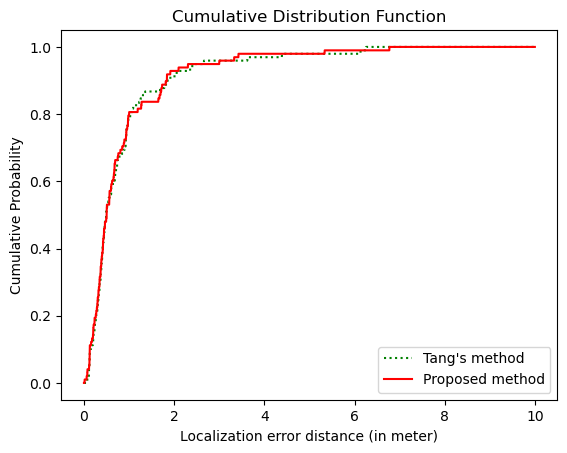}
        \caption{CDF comparison with Tang's method}
        \label{CDF comparison tang}
    \end{subfigure}
    \caption{CDF comparison with existing fusion ideas}
    \label{CDF4}
\end{figure*}

Fig.\ref{CDF comparison-gamma} compares the proposed method with the probability-threshold data fusion. The threshold $\gamma$ = 0.9 has the worst performance in error interval [0, 4], because $\gamma$ = 0.9 is not sufficient to include the important candidate positions, in this case, the true positions are likely to be excluded from the candidate set. One can note that increasing the value of $\gamma$ can improve the performances, but it is not very helpful in limiting the large errors. In particular, $\gamma$ = 0.98 has the worst performance in limiting large errors bigger than 4 meters, because a big $\gamma$ value means relaxed constraint. $\gamma$ = 1 represent no constraint. The proposed method shows its advantage in limiting the large localization errors if error is bigger than 2 meters.

The proposed method is compared with Top-$K$ data fusion in Fig.\ref{CDF comparison-K}. For the data fusion with $K = 5$, it has the worst performance in error interval [0, 3.5], because it is likely to exclude the true position from the candidate set. Top-$K$ with $K = 10$ can improve the performance in error interval [0, 3.5], but it is not good at limiting big errors. $K = 15$ brings even higher risk of big localization error than $K=10$. Similarly, a big $K$ means relaxed threshold.

Fig.\ref{CDF comparison tang} provides a comparison between the proposed method and 'Tang's method' \cite{Tang}, which is a recently published hybrid data fusion criterion. The proposed method and Tang's method have very closed performances. The detailed comparison is given in Fig. \ref{RMSE comparison}.

Fig. \ref{RMSE comparison} compares their localization errors in terms of RMSE, MAE and STD, respectively. One can note that the RMSE and MAE of the proposed method is 1.32 meters and 1.06 meters, respectively, which are smaller than those of other methods. Thus the proposed method can achieve the highest localization accuracy. In addition, the proposed method has the smallest STD value (1.02 meters), which means that its localization performance is more stable.

\begin{figure}[h]
\centering
\includegraphics[width=3.5in]{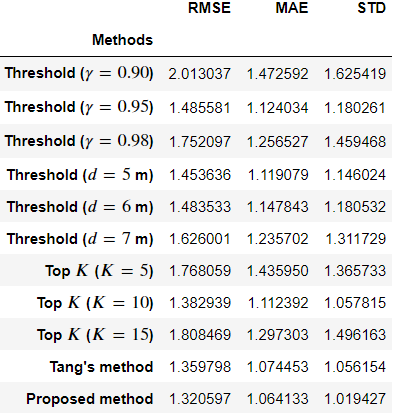} 
\caption{Localization performance comparison (RMSE, MAE and STD)}
\label{RMSE comparison}
\end{figure}

\section{Conclusions}
This paper proposes a soft WiFi-Visual data fusion method for indoor robot localization. The localization problem is modeled as a classification. The WiFi features are represented in image form in order to fuse the WiFi features and visual features. The fused WiFi-Visual features are jointly exploited by a classification neural network to produce a likelihood vector, which can classify the input features to the most probable position. The visual features and WiFi features are also separately exploited by neural networks to produce another two likelihood vectors. The three likelihood vectors are fused by Hadamard product and median filtering to produce the final likelihood vector. The proposed method is tested on an indoor robot (TIAGO++). The test shows that the proposed method can use 10 WiFi samples and 4 low-resolution images ($58 \times 58$ in pixels) to localize the robot with an average error distance about 1.32 meters. The experiment test is 3 months after the data collection in a general teaching building, whose WiFi and visual environments are partially changed. This indirectly shows the robustness of the proposed method. This localization mode can initialize other source-based localization operating in a non-optimal manner in the event of a robot kidnapping, such as kidnapped LiDAR.

\section*{Acknowledgment}
The authors would like to thank Ilinykh Arthur for his participation in our project as an engineering student by programming the robot to take photo sequence.
}

\end{document}